\documentclass[10pt,twocolumn,letterpaper]{article}

\usepackage{cvpr}              
\usepackage{multirow}

\newcommand{\figref}[1]{Fig.~\ref{#1}}
\newcommand{\tabref}[1]{Tab.~\ref{#1}}
\newcommand{\secref}[1]{Sec.~\ref{#1}}

\usepackage[table]{xcolor}
\definecolor{colorfirst}{RGB}{255, 204, 204}
\definecolor{colorsecond}{RGB}{255, 230, 204}
\definecolor{colorthird}{RGB}{255, 251, 214}
\newcommand{\cellfirst}{{\cellcolor{colorfirst}}}
\newcommand{\cellsecond}{{}}
\newcommand{\cellthird}{{}}

\definecolor{cvprblue}{rgb}{0.21,0.49,0.74}
\usepackage[pagebackref,breaklinks,colorlinks,allcolors=cvprblue]{hyperref}

\def\paperID{10199} 
\def\confName{CVPR}
\def\confYear{2025}

\title{Fine-Grained Controllable Apparel Showcase Image Generation via Garment-Centric Outpainting}
\author{
    Rong Zhang\textsuperscript{1} \quad
    Jingnan Wang\textsuperscript{1} \quad
    Zhiwen Zuo\textsuperscript{1} \quad
    Jianfeng Dong\textsuperscript{1} \quad \\
    Wei Li\textsuperscript{2} \quad
    Chi Wang\textsuperscript{3} \quad
    Weiwei Xu\textsuperscript{3} \quad
    Xun Wang\textsuperscript{1} \\
    \textsuperscript{1}Zhejiang Gongshang University \quad
    \textsuperscript{2}Inceptio \quad
    \textsuperscript{3}Zhejiang University
}

\begin{document}

\twocolumn[
{
\renewcommand\twocolumn[1][]{#1}
\maketitle
     \centering
        \includegraphics[width=0.96\linewidth]{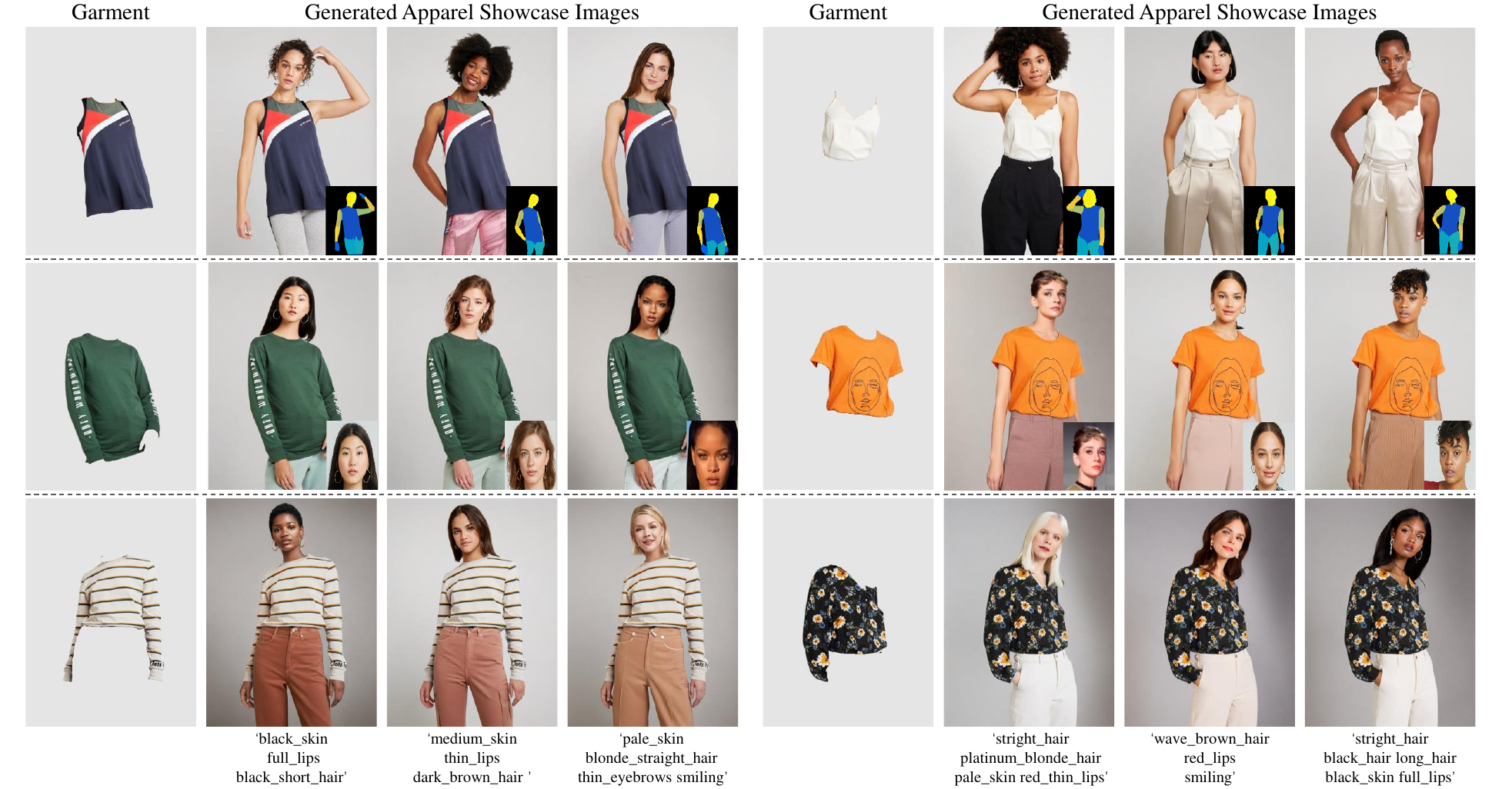}
        \captionof{figure}{
        Given a garment image segmented from a dressed mannequin or a person, our method can generate apparel showcase images via garment-centric outpainting under the guidance of face images or fine-grained text attributes. From top to bottom: (1) The results conditioned on the automatically generated diverse pose maps. (2) The results conditioned on different face images. (3) The results conditioned on fine-grained text prompts.
        }
        \vspace{2em}
        \label{fig:teaser}
}
]

\begin{abstract}
In this paper, we propose a novel garment-centric outpainting (GCO) framework based on the latent diffusion model (LDM) for fine-grained controllable apparel showcase image generation. The proposed framework aims at customizing a fashion model wearing a given garment via text prompts and facial images. Different from existing methods, our framework takes a garment image segmented from a dressed mannequin or a person as the input, eliminating the need for learning cloth deformation and ensuring faithful preservation of garment details. The proposed framework consists of two stages. In the first stage, we introduce a garment-adaptive pose prediction model that generates diverse poses given the garment. Then, in the next stage, we generate apparel showcase images, conditioned on the garment and the predicted poses, along with specified text prompts and facial images. Notably, a multi-scale appearance customization module (MS-ACM) is designed to allow both overall and fine-grained text-based control over the generated model's appearance. 
Moreover, we leverage a lightweight feature fusion operation without introducing any extra encoders or modules to integrate multiple conditions, which is more efficient.
Extensive experiments validate the superior performance of our framework compared to state-of-the-art methods. 
\end{abstract}
\vspace{-0.4in}    
\section{Introduction}
\label{sec:intro}

\begin{figure}[t]
  \centering
  \includegraphics[width=\linewidth]{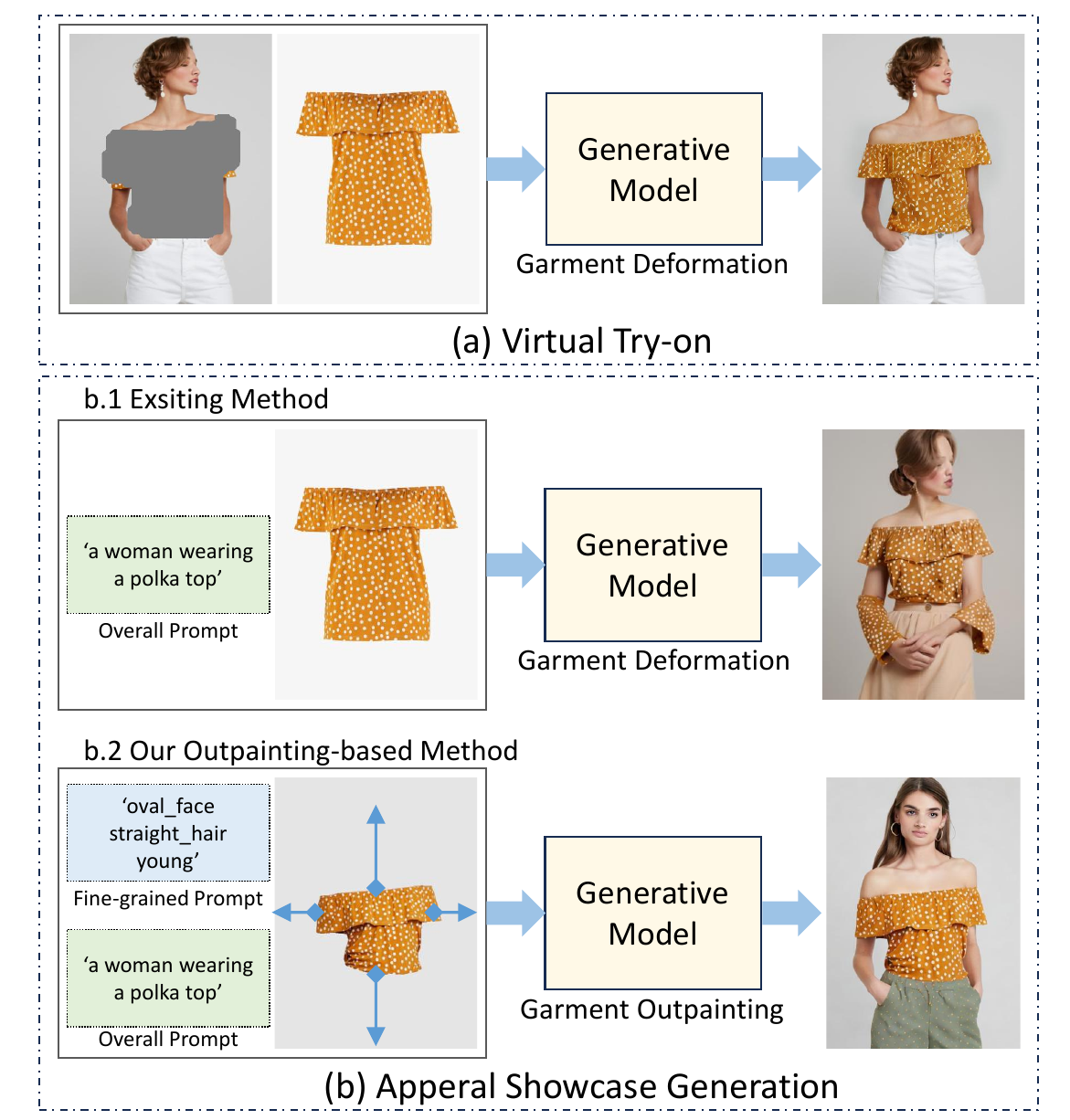}
   \vspace{-2em}
   \caption{Task comparison of the existing fashion-related image generation methods. (a): virtual try-on based method; (b): the showcase generation method, where b.1 is the existing method and b.2 is our outpainting-based method. Note that our method can preserve the garment details better and enable fine-grained text prompts for customization. }
   \label{fig:vs}
\end{figure}

The rapid advances in AI-generated content (AIGC) have been revolutionizing numerous industries, with the fashion sector being one of the most profoundly impacted. Existing research in the fashion domain has primarily focused on image-based virtual try-on (VTON)~\cite{han2018viton,han2019clothflow,kim2024stableviton}, which can enhance the user experience of online shoppers by visualizing the try-on effect of the apparel. As shown in Fig.~\ref{fig:vs} (a), the core of VTON lies in learning to accurately deform an apparel image to fit a given cloth-agnostic person image and faithfully preserving the apparel's details in the generated results, which is essentially an instance-guided image inpainting task. 


Recently, there emerged another line of work that emphasizes customizing the fashion models wearing the given garment images as depicted in Fig.~\ref{fig:vs} (b), dubbed as apparel showcase image generation.  
The goal of this task is to automatically create realistic and controllable online garment showcase advertisement images as those present in e-commerce platforms, which has great potential to freely customize the fashion models' characteristics and reduce the operational costs for online apparel retailers. Compared to VTON, the apparel showcase generation task also needs to faithfully preserve the garment details. Yet, it presents an additional challenge of ensuring fine-grained controllability over the generated fashion models.

In recent years, with the tremendous success of latent diffusion models (LDMs)~\cite{LDM} in text-to-image generation, some apparel showcase generation approaches~\cite{chen2024magic,huang2024parts} have attempted to build frameworks based on them. Magic Clothing (MC)~\cite{chen2024magic} may be the first work in this area, enabling customizing the fashion models wearing the given garment via diverse text prompts. MC introduces a garment encoder to extract the garment features and inject them into LDMs. 
It also utilizes overall text annotations to allow coarse control over the synthesized model's appearance. 
More recently, a controllable human image synthesis framework called Parts2Whole~\cite{huang2024parts} is proposed to synthesize human images using multiple parts of human appearance, such as faces, hairs, and upper and lower body clothes, which can be applied for apparel showcase generation as well. Similar to MC, Parts2Whole designs an appearance encoder to obtain hierarchical appearance features of multiple human parts.
Although these approaches have achieved promising results, we argue that they still struggle with the aforementioned two key challenges of the apparel showcase generation task: 1) the faithful preservation of the garment details, and 2) the fine-grained controllability over the synthesized model's appearance.
First, as shown in Fig.~\ref{fig:vs} (b.1), their frameworks cannot get rid of learning to deform the cloth, which may result in cloth distortions or detail loss in the generated garment.
Such defects can seriously limit the practical usage of existing methods in real-world e-commerce scenarios. Second, existing methods may fail to control the detailed attributes of the synthesized fashion models, \textit{e.g.,} hairstyle, and skin colors, due to the lack of specific designs.  

To address these limitations, we propose a novel fine-grained controllable apparel showcase image generation framework via garment-centric outpainting (GCO). It outpaints an already-deformed garment image to customize a model wearing it as shown in ~\figref{fig:vs} (b.2). Such deformed cloth images can be easily obtained by dressing a mannequin or having a person wear the clothes. Therefore, our method eliminates the need for garment deformation, avoiding potential cloth distortions and ensuring faithful preservation of garment details in the synthesized results. 
Also, as a large number of clothed person images are available on the internet, our framework can be trained using these images without the necessity of collecting paired data. 

The proposed framework mainly comprises two stages. In the first stage, we introduce a garment-adaptive pose prediction model, which trains a diffusion model from scratch to generate diverse poses that fit the given deformed garment. In the second stage, we condition the LDM on the garment and the predicted poses, along with specified text prompts and facial images, to generate the apparel showcase images. To empower our framework with the capability to control both the overall and fine-grained characteristics of the synthesized fashion models, we further design a multi-scale appearance customization module (MS-ACM). MS-ACM handles two types of text prompts through a Showcase BLIP and a Face BLIP~\cite{li2022blip}, respectively: one to control the overall appearance of the showcase images, and the other to adjust its detailed facial features.
In addition, we leverage a lightweight feature fusion operation for incorporating multiple conditions into LDMs without introducing any additional encoders or modules, which improves efficiency compared to existing commonly adopted LDM condition methods~\cite{zhang2023adding,ye2023ip}.
Equipped with these careful designs, the proposed framework, GCO, demonstrates higher quality and controllability in apparel showcase generation. We conduct extensive experiments to validate the superior performance of GCO over the state-of-the-art methods.

To summarize, our key contributions are as follows:
\begin{itemize}
    \item To the best of our knowledge, we for the first time propose a fine-grained controllable apparel showcase generation framework based on garment-centric outpainting, eliminating the need for cloth deformation and ensuring faithful preservation of garment details.
    \item We design a multi-scale appearance customization module (MS-ACM), allowing both overall and fine-grained control over the generated model's appearance. Moreover, MS-ACM is a plug-in module that can be easily combined with other related methods.
    \item Qualitative and quantitative experiments show that our framework consistently outperforms state-of-the-art methods in both realism and controllability of the results. 
\end{itemize}


\section{Related Works}

\subsection{Diffusion Models}
Diffusion models~\cite{sohl2015deep,ho2020denoising,song2019generative,song2020score} have attracted significant research attention recently, as they have beat generative adversarial networks (GANs)~\cite{goodfellow2014generative} on image synthesis~\cite{dhariwal2021diffusion} and achieved great success in text-to-image generation~\cite{LDM,ramesh2022hierarchical,saharia2022photorealistic}. However, the inference speed of diffusion models is relatively slow, as they require simulating a Markov chain for many steps in order to produce a sample. To accelerate inference, denoising diffusion implicit models (DDIM)~\cite{song2020denoising} and pseudo numerical methods for diffusion models (PNDM)~\cite{liu2022pseudo} introduce new noise schedulers for diffusion models. Further, latent diffusion models (LDMs)~\cite{LDM} propose to build diffusion models in the latent space of a pre-trained variational autoencoder (VAE)~\cite{kingma2013auto,rezende2014stochastic}, which achieves a balance between computational efficiency and image quality.

\subsection{Image-Based Virtual Try-on}
Image-based virtual try-on (VITON) task is designed to generate realistic try-on images based on the given garment image and the reference person image.
Most prior VITON methods consist of two main phases: 1) the cloth warping phase, and 2) the try-on generation phase. In the first phase, thin plate spline (TPS)~\cite{duchon1977splines} and appearance flow~\cite{li2019dense} are commonly adopted for cloth warping. In the second phase, GANs~\cite{goodfellow2014generative} often play a pivotal role in refining try-on results. Recently, many researchers have developed methods based on LDMs~\cite{LDM} due to their superior image quality. Some approaches~\cite{gou2023taming,li2023virtual,wang2024fldm} still follow the previous two-phase framework, substituting GANs with LDMs in the second phase for synthesizing more realistic try-on results, while others~\cite{morelli2023ladi,zhu2023tryondiffusion,yang2024texture,choi2024improving,kim2024stableviton,zeng2024cat} build end-to-end frameworks by introducing garment encoders into LDMs and then utilizing attention modules to perform implicit garment warping. Nonetheless, few works have paid attention to apparel showcase image generation. Magic Clothing (MC)~\cite{chen2024magic} is the most related work to ours in this area, which customizes characters wearing the target garment with diverse text prompts. However, compared to our method, MC falls short in cloth details and lacks precise control over the fine-grained attributes of the generated model's appearance.

\begin{figure*}[t]
  \centering
  \includegraphics[width=\linewidth]{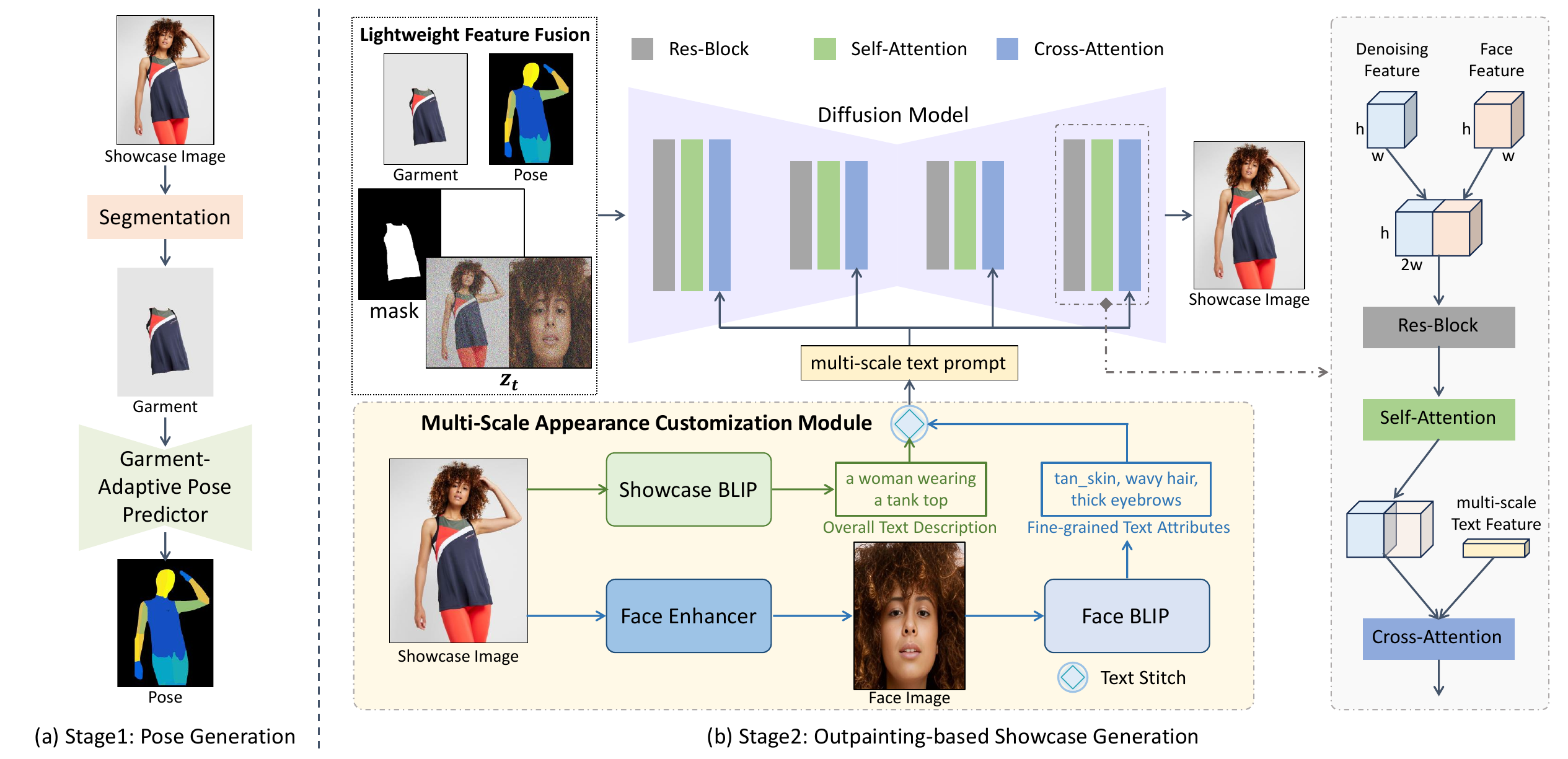}
   \caption{The overview of our GCO framework. (a): The pipeline of the garment-adaptive pose predictor. In the inference stage, we can utilize it to sample diverse pose maps that fit a target garment. (b): The training process of the outpainting-based showcase generation stage. It consists of a multi-scale appearance customization module and a lightweight feature fusion operation. For the multi-scale appearance customization module, our method extracts the overall image descriptions as coarse conditions and the fine-grained face attributes as detailed conditions through two different BLIP models. For the lightweight feature fusion, we fuse the multiple input conditions including the garment, the pose, the mask, and the facial features with the image to be generated through spatial or channel-wise concatenation. 
   After training, our method can take facial images, overall showcase descriptions, or fine-grained text attributes as optional inputs to generate diverse showcase images.}
   \label{fig:pipeline}
\end{figure*}

\subsection{Human Image Generation}
Human image generation (HIG) task aims at synthesizing realistic and diverse human images.
We roughly categorize HIG into transfer-based and synthesis-based methods. Given source human images and target pose conditions, transfer-based algorithms~\cite{ma2017pose,han2023controllable,shen2023advancing,lu2024coarse} are expected to output photorealistic images with source appearance and target poses. In contrast, synthesis-based HIG methods~\cite{fruhstuck2022insetgan,fu2022stylegan,yang20233dhumangan,ju2023humansd,liu2023hyperhuman,wang2024towards,zhu2024mole,huang2024parts} concentrate on synthesizing high-quality human images conditioned on poses, text prompts, or faces. Our work is more related to the latter one. 
Early works on synthesis-based HIG are mainly based on GANs~\cite{fruhstuck2022insetgan,fu2022stylegan,yang20233dhumangan}, while many recent approaches have delved into designing specialized frameworks based on LDMs for better quality and controllability, such as incorporating more annotations~\cite{liu2023hyperhuman}, designing losses based on human-centric priors~\cite{ju2023humansd,wang2024towards}, or utilizing multiple human-part images~\cite{huang2024parts,zhu2024mole}. Particularly, Parts2Whole~\cite{huang2024parts} proposes to synthesize human images based on multiple human parts, including faces and upper-body clothes.
Aside from these methods, a lot of personalized text-to-image methods~\cite{ye2023ip,li2024photomaker,shi2024instantbooth,peng2024portraitbooth} have been proposed to customize portraits conditioned on facial images.
Different from existing HIG works, we focus on customizing fashion models wearing the given garment in this paper, with emphasis on faithful preservation of the garment details and fine-grained control over the generated model's appearance.


\section{Approach}

In this section, we present GCO, a fine-grained controllable garment-centric outpainting framework for apparel showcase image generation. 
As shown in ~\figref{fig:pipeline}, the proposed framework consists of two stages. In the first stage, given a garment image segmented from a dressed mannequin or a person, a garment-adaptive pose predictor is introduced to generate the corresponding pose that fits it. In the second stage, we build a garment-centric outpainting model based on latent diffusion models (LDMs)~\cite{LDM}, taking the garment and the predicted pose along with text prompts and facial images as the input, to customize the fashion models wearing the given garment.
In what follows, ~\secref{sec:pre} briefly reviews the LDMs, which are the basis of the garment-centric outpainting model. ~\secref{sec:pose} and ~\secref{sec:face} present the details of the proposed garment-adaptive pose predictor and the garment-centric outpainting model, respectively. 




%

\subsection{Preliminaries}\label{sec:pre}

LDM is a type of diffusion models built on the latent space of a variational autoencoder (VAE). It consists of two main components: a VAE and a denoising UNet $\epsilon_\theta$. The VAE maps images between the pixel space and the low-dimensional latent feature space, while the denoising UNet $\epsilon_\theta$ is responsible for generating latent encodings of images through a denoising process. The training of the denoising UNet $\epsilon_\theta$ is the same as in the Denoising Diffusion Probabilistic Model (DDPM)~\cite{ho2020denoising}, where a forward diffusion process and a backward denoising process of $T=1000$ steps are formulated. In the forward diffusion process, random Gaussian noise $\epsilon$ is progressively added to the image latent $z_0$:
\begin{equation}
z_t = \sqrt{\bar{\alpha}_t}z_0+\sqrt{1-\bar{\alpha}_t} \epsilon, \quad t\in\{1,2,...,T\},
\end{equation}
 where $\{\ \bar{\alpha}_1, \bar{\alpha}_2, ..., \bar{\alpha}_T\}$ are derived from a fixed variance schedule. The denoising process learns to predict the noise added through $\epsilon_\theta(z_t,t,c)$, where $c$ is the textual embedding extracted by a CLIP~\cite{radford2021learning} text encoder. The optimization is performed by minimizing the following objective:
\begin{equation}
\mathcal{L}_{SD} = \mathbb{E}_{z,\epsilon\sim\mathcal{N}(0,1),t} [ || \epsilon - \epsilon_{\theta} (z_t,t,c) ||_2^2].
\end{equation}


\subsection{Garment-Adaptive Pose Predictor}\label{sec:pose}
Without any human structural priors, directly outpainting a deformed garment to synthesize a fashion model wearing it is very difficult. Therefore, we first design a garment-adaptive pose prediction model to generate the corresponding poses that fit the given garment.  
Specifically, the garment-adaptive pose predictor is based on the LDM framework. Given an input apparel showcase image $I \in \mathbb{R}^{H\times W \times 3}$, we obtain the deformed garment image $C\in \mathbb{R}^{H\times W\times 3}$ and the corresponding pose map $P \in \mathbb{R}^{H\times W\times 3}$ from it by a pre-trained cloth segmentation model~\cite{dabhi2021clothsegmentation} and a DensePose model~\cite{guler2018densepose}, respectively. We then train a UNet denoiser $\epsilon_{\theta_p}$ from scratch to generate the pose $P$ conditioned on the garment $C$. Since $P$ and $C$ are spatially aligned, we simply concatenate the garment image's latent encoding $z_c$ with the pose image's noised latent encoding  ${z_p}_t$ in the channel dimension as the input to the UNet denoiser $\epsilon_{\theta_p}$. The objective function for training is as follows:
\begin{equation}
\mathcal{L}_{pose} = \mathbb{E}_{p, c,\epsilon\sim\mathcal{N}(0,1),t} [ || \epsilon - \epsilon_{\theta_p} ({z_p}_t, z_c, t) ||_2^2].
\end{equation}

During the inference, thanks to the stochasticity of the reverse denoising process, we can utilize the garment-adaptive pose predictor to sample diverse pose maps that fit the given garment.

\subsection{Garment-Centric Outpainting Model}\label{sec:face}
In the second stage, we propose a garment-centric outpainting model~(\figref{fig:pipeline} (b)) based on LDM to generate apparel showcase images conditioned on the given garment and the predicted pose maps, along with the text prompts and the facial images. We meticulously design a lightweight feature fusion operation and a multi-scale appearance customization module, to achieve fine-grained control over the synthesized models' appearance with an efficient framework.


\paragraph{Lightweight Feature Fusion.}
In this stage, we have two kinds of input conditions~(\figref{fig:pipeline} (b)): 1) conditions that are spatially aligned with the target showcase image $I$, including the garment image $C$, the pose map $P$, and the garment mask $M \in \mathbb{R}^{H\times W}$ indicating the regions to be outpainted, and 2) conditions that are not spatially aligned with the target showcase image $I$, \ie the user-specified facial images $F$ of the synthesized fashion models.
A naive way to incorporate these conditions into LDM is by using the combination of ControlNet~\cite{zhang2023adding} and IP-Adapter~\cite{ye2023ip}, dubbed as ``ControlNet+IP-Adapter" for convenience, where ControlNet is suitable for spatially aligned conditions, while IP-Adapter provides attention-based feature injection for the input facial image.
However, the above approach introduces additional image encoders and attention modules into LDM, resulting in huge computational and memory overhead.
In order to build an efficient model, we design a simple yet effective lightweight feature fusion operation almost without any changes to the LDM.
For the conditions that are spatially aligned with $I$, we fuse them through channel-wise concatenation. All we need to do is to modify the first conv layer in the UNet denoiser to support the input with more channels.
For the input facial image $F$ that is not spatially aligned with $I$, we concatenate $F$ with $I$ in the spatial dimension, relying on the self-attention modules within LDM to decide which facial features of
$F$ the denoising features require to attend to. 
Compared to ``ControlNet+IP-Adapter", our method only adds negligible number of parameters to the first conv layer of the LDM, significantly reducing the computational and memory overhead.



\paragraph{Multi-scale Appearance Customization Module.}


The fine-grained characteristics of the synthesized models, like the detailed facial attributes, are the most important appearance to be controlled; nonetheless, existing methods struggle to control them effectively due to the lack of specific designs.
To tackle the problem, we present a multi-scale appearance customization module (MS-ACM) that allows both overall and fine-grained control over the generated model's appearance through two types of text prompts, respectively. Specifically, we caption the showcase image $I$ with the original BLIP~\cite{li2022blip}, dubbed Showcase BLIP, to generate the overall coarse text prompt $T_c$. Meanwhile, we take the detailed textual attributes of the facial region of $I$ as the fine-grained text prompts. To this end, we first design a face enhancer module to obtain the high-quality facial image of $I$. In the face enhancer module, we first adopt Yolov5~\cite{ultralytics2021yolov5} to detect the face region of $I$. Then a face restoration model CodeFormer~\cite{zhou2022codeformer} and a super-resolution model RealESRGAN~\cite{wang2021real} are used in a sequential manner to obtain a high-quality facial image $F$. 
Since existing fashion datasets are short of detailed facial descriptions of the fashion models, we further fine-tune a BLIP model called Face BLIP on the CelebA dataset~\cite{liu2015faceattributes}, which contains fine-grained facial attributes. After fine-tuning, we keep the Face BLIP model frozen during both training and inference to generate fine-grained textual descriptions $T_f$ of the given facial image. Finally, $T_c$ and $T_f$ are stitched together to a multi-scale text prompt and input to the denoising UNet through cross-attention blocks.



\begin{figure*}[t]
  \centering
  \includegraphics[width=\linewidth]{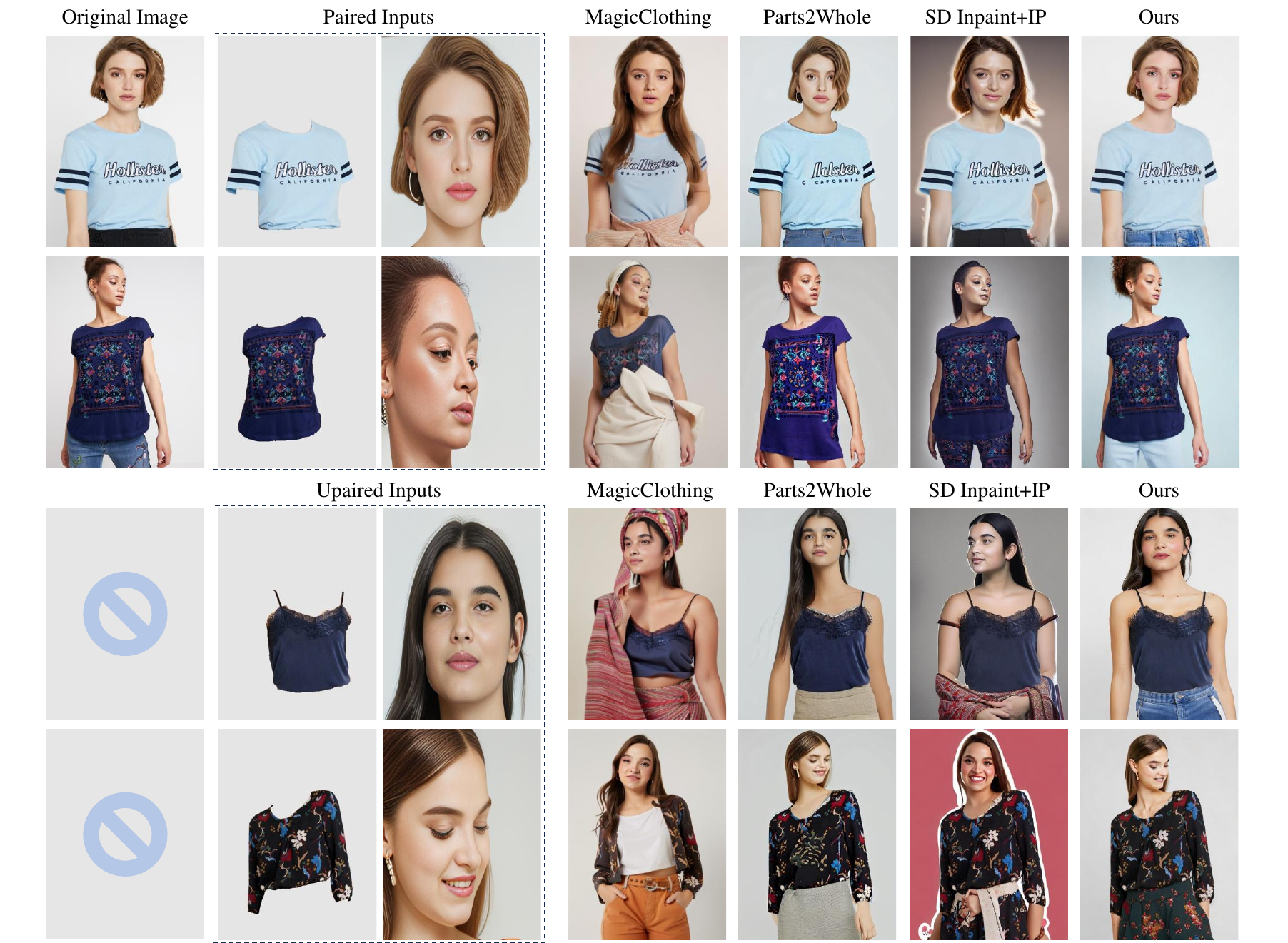}
   \caption{Comparison with the baseline methods. The first two rows show paired results where the input garment and face are from the same original showcase image (The first column). The last two rows show unpaired results where the input garment and face are from different images.}
   \label{fig:results}
\end{figure*}



\section{Experiments}

\subsection{Implementation Details}
Our method focuses on the task of apparel showcase generation. 
By adopting an outpainting-based framework, it does not require paired data of well-posed front-view garment images and corresponding showcase images. 
To evaluate GCO and other related methods, we conduct experiments on the virtual try-on benchmark VITON-HD, which contains $14221$ training images and $2032$ testing images. 
Our method builds on the Stable Diffusion 1.4~\cite{LDM}.
All experiments are implemented with Pytorch and performed on an NVIDIA A100 80G GPU.



\begin{table*}[t]
\centering
\caption{Quantitative comparison with different methods.}
\label{tab:comparison}
\begin{tabular}{cccccccc} 
\toprule
Method        & MP-LPIPS↓ & Clo-SSIM↑ & Face-LPIPS↓ & FID↓  & LPIPS↓ & DreamSim↓ & CLIP-i↑  \\ 
\midrule
SD Inpaint    &    \cellthird 0.14      & \cellsecond 0.96      &     -        & 66.56 & 0.65   & 0.34      & 0.77     \\
SD Inpaint+IP & \cellthird 0.14      & \cellthird 0.95      & 0.465       & 79.52 & 0.57   & \cellthird 0.32      & 0.80      \\
Parts2Whole   & 0.07      & 0.94      & \cellthird 0.379       & \cellsecond 25.65 & \cellsecond 0.31   & \cellsecond 0.20      & \cellsecond 0.89     \\
MagicClothing & \cellsecond 0.20      & 0.87      & \cellsecond 0.360       & \cellthird 57.38 & \cellthird 0.42   & 0.37      & \cellthird 0.82     \\
Ours          & \cellfirst 0.045     & \cellfirst 0.97      & \cellfirst 0.127       & \cellfirst 11.36 & \cellfirst 0.15  & \cellfirst 0.09      & \cellfirst 0.95     \\
\bottomrule
\end{tabular}
\end{table*}

\begin{figure}[t]
  \centering
  \includegraphics[width=\linewidth]{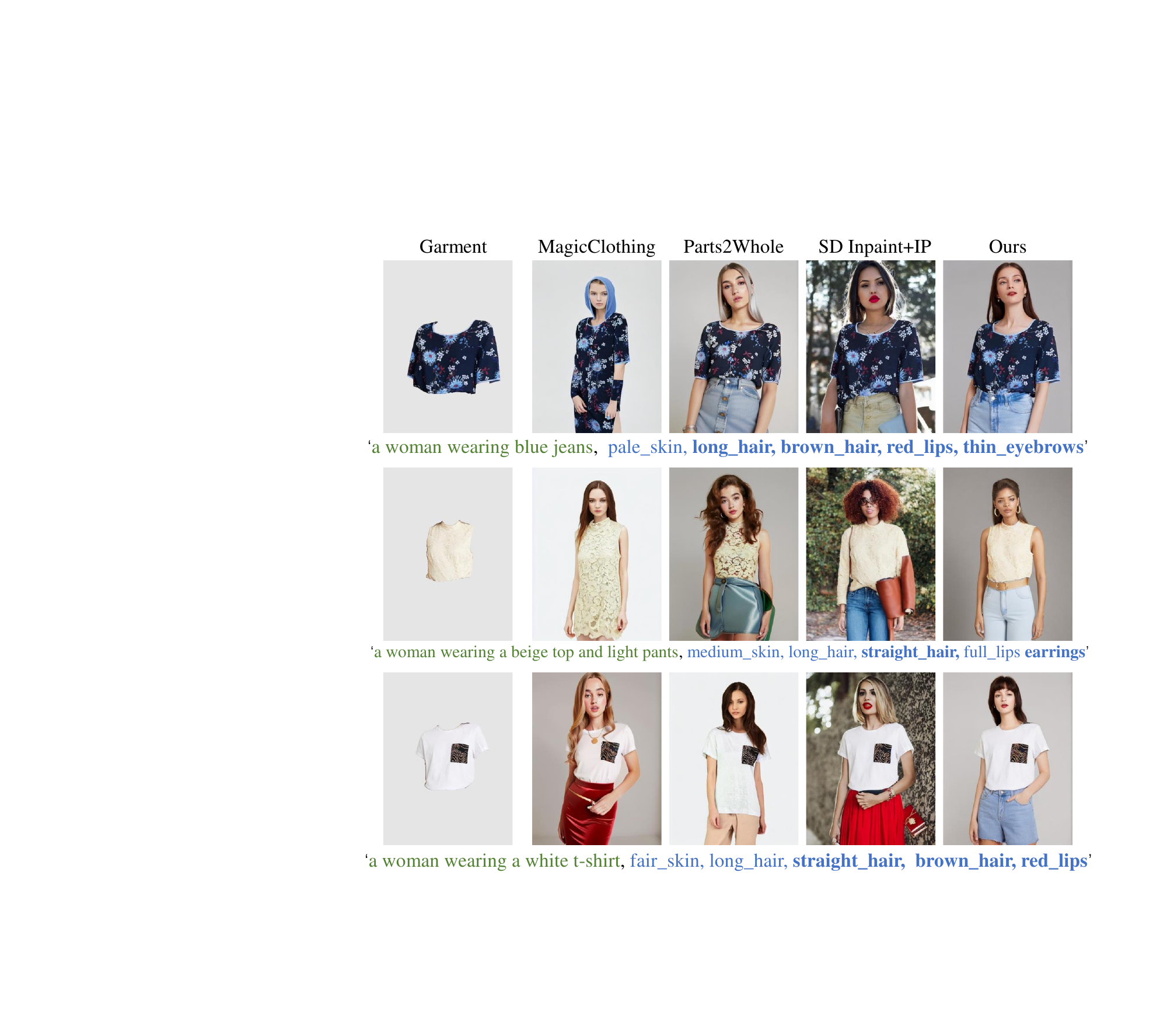}
   \caption{Comparison of multi-scale appearance customization module. The overall text prompt is marked in green while the fine-grained attributes are marked in blue. The attributes that may be mismatched are highlighted in bold.}
   \label{fig:text-compare}
\end{figure}

\subsection{Comparisons}

\paragraph{Baselines.} As there are few methods targeted at apparel showcase image generation, we compare GCO with the related inpainting-based and garment-driven image synthesis approaches to evaluate the performance of our method. The baselines are as follows:

\begin{itemize}
    \item SD Inpaint is the Stable Diffusion model trained for the image inpainting with pose conditions added through the ControlNet-OpenPose~\cite{zhang2023adding}. 
    \item SD Inpaint+IP is the above SD Inpaint model with an additional IP-Adapter~\cite{ye2023ip} to customize facial appearance.
    \item Parts2Whole~\cite{huang2024parts} is a framework designed for generating portraits from multiple reference images, including pose images and various aspects of human appearance. 
    \item MagicClothing~\cite{chen2024magic} is an LDM-based garment-driven image synthesis network. We use the well-posed front-view garment images as its inputs.
\end{itemize}
For Parts2Whole, we re-trained the model on the VITON-HD dataset for comparison. For other methods, official pre-trained models are utilized to produce results.

\paragraph{Metrics.}

To evaluate the methods, we calculate the metrics in a paired setting where the input target garment and face image are extracted from the same ground truth showcase image. Learned Perceptual Image Patch Similarity (LPIPS)~\cite{zhang2018unreasonable}, Frechet Inception Distance (FID)~\cite{heusel2017gans} and DreamSim~\cite{fu2023dreamsim} are utilized to measure the realism of the generated images from different dimensions. Besides, we adopt the structural similarity of the cloth region (Clo-SSIM) and the Matched-Points-LPIPS (MP-LPIPS)~\cite{chen2024magic} to evaluate whether the characteristics of the target garment are well-preserved. We also calculate the LPIPS of the generated face region (Face-LPIPS) to measure the perceptual quality of the face. In addition, we use CLIP-i to estimate the similarity of the CLIP space between the generated images and the ground truths. 



\paragraph{Results and Analysis.}

\figref{fig:results} demonstrates the qualitative comparisons between GCO and state-of-the-art baseline methods. The first two rows show the results with the input garment and face from the same original showcase image, while the last two rows present the results with the input garment and face from different showcase images. We observe that our method achieves better results with fewer artifacts. On one hand, as GCO utilizes a deformation-free outpainting-based framework, the garment details, especially the characters and patterns, are better preserved. In contrast, the garment deformation in baseline methods may introduce undesired distortions or color deviation,~\eg the cloth in the 4th row, 4th column has been changed from a blouse to a coat. On the other hand, GCO can maintain facial identity and structural integrity more accurately via lightweight feature fusion. Meanwhile, the compared methods are unable to preserve the facial characteristics.

\tabref{tab:comparison} presents the quantitative comparison results between GCO and the baseline methods. Our method outperforms others on all the metrics. Clo-SSIM and MP-LPIPS  numerically validate the effectiveness that GCO can preserve the garment details. Face-LPIPS further demonstrates that our method can maintain the identity of the given face images. Other metrics present the superiority of our method in terms of image quality and realism. Note that since the SD Inpaint accepts no facial image as input, we did not calculate its Face-LPIPS.

In addition, we conduct experiments to verify the effectiveness of these methods in fine-grained appearance customization.~\figref{fig:text-compare} shows the comparisons under the guidance of fine-grained text attributes. The baseline methods tend to ignore some of the attributes such as `straight hair',  `thin eyebrows', and `red lips'. In contrast, GCO is capable of maintaining most of the desired details.

\begin{figure*}[t]
  \centering
  \includegraphics[width=0.95\linewidth]{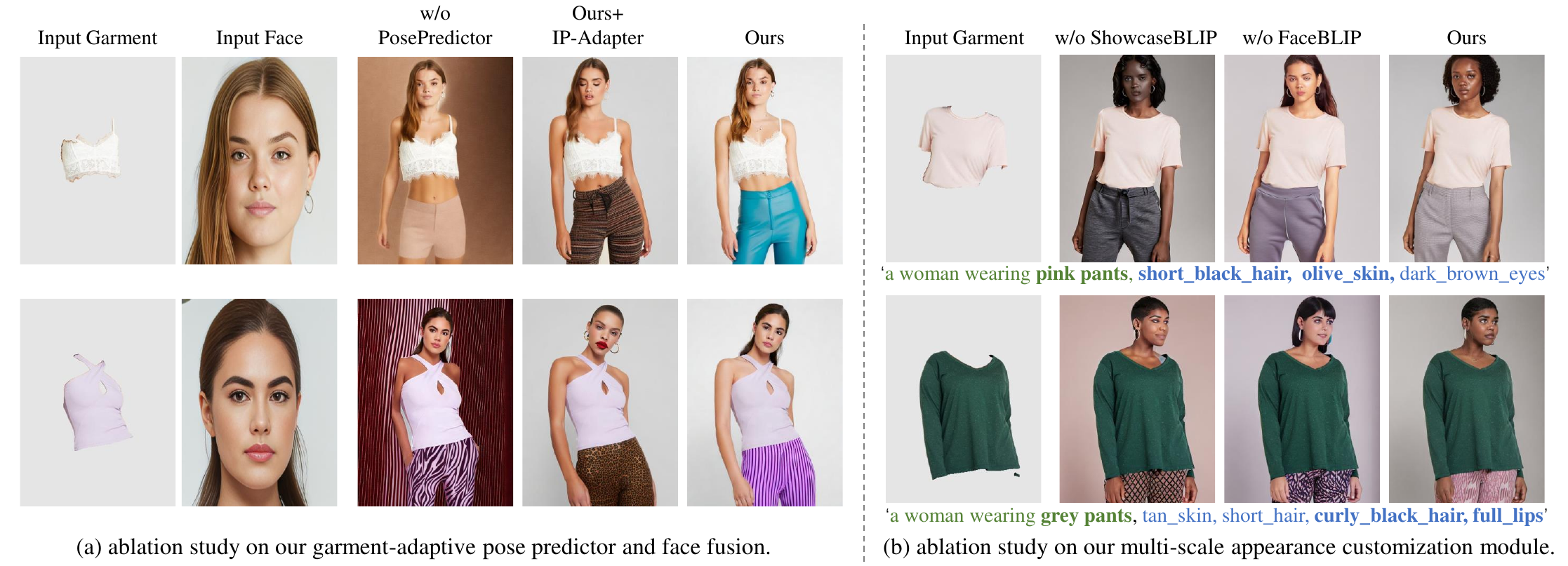}
   \caption{Results of ablation study. (a): ablation study on garment-adaptive pose predictor and face fusion. The third column shows the results without the pose predictor. The fourth column shows the results of replacing our face fusion operation with an IP-Adapter. (b): ablation study on multi-scale appearance customization module. The overall text prompt is marked in green while the fine-grained attributes are marked in blue. The attributes that may be mismatched are highlighted in bold.}
   \vspace{-1em}
   \label{fig:ablation}
\end{figure*}

\begin{table}[t]
\centering
\caption{User study.}
\label{tab:userstudy}
\begin{tabular}{cccc}
\toprule
 Method     & Face & Garment & Overall \\
\midrule
MagicClothing & \cellsecond 18.26\%   & \cellsecond 24.35\%  & \cellsecond 24.46\%      \\
Parts2Whole   & \cellthird 12.39\%   & \cellthird 8.70\%   & \cellthird 7.39\%       \\
SD Inpaint+IP    & 6.20\%    & 8.04\%   & 6.52\%       \\
Ours  & \cellfirst 63.15\%   & \cellfirst 58.91\%  & \cellfirst 61.63\%    \\ 
\bottomrule
\end{tabular}
\end{table}

\paragraph{User Study.}
We conduct a user study to further evaluate perceptual realism. 
Specifically, We randomly select $20$ target garments from the test set. For each garment, we generate the showcase images using a reference face image. 
We demonstrate the target garment, face image, and the generated results from all methods simultaneously to the users. Then we ask them to pick $3$ kinds of the most qualified result: (1) Face: the result that best preserves the face identity; (2) Garment: the result that best matches the target garment image; (3) Overall: the result with the highest overall image quality. $46$ participants are recruited to finish the human perceptual study. More than $900$ samples are collected.
As the statistical results in \tabref{tab:userstudy} show, our method gains the highest user preference across all the dimensions, outperforming the second place by a large margin.


\begin{table}[t]
\centering
\caption{Ablation Study. }
\vspace{-8pt}
\label{tab:ablation}
\resizebox{\linewidth}{!}{
\begin{tabular}{cccccc} 
\toprule
Method            & Face-LPIPS↓             & FID↓                   & LPIPS↓                 & DreamSim↓               & CLIP-i↑                 \\ 
\midrule
w/o PosePredictor & 0.5472                        &         31.43               &      0.208                  &     0.187                    &   0.866                      \\
w/o ShowcaseBLIP  & \cellsecond 0.1272      & 30.08                  & 0.165                  & 0.117                  & 0.950                   \\
w/o FaceBLIP      & \cellthird 0.1282                  & \cellsecond 12.79      & \cellsecond 0.157      & \cellthird 0.098                  & 0.949                   \\
Ours+IP-Adapter   & 0.4709                  & 24.34                  & 0.256                  & 0.194                  & 0.887                   \\
Ours              & \cellfirst 0.1266 & \cellfirst 11.36 & \cellfirst 0.148 & \cellfirst 0.090 & \cellfirst 0.952                   \\
\bottomrule
\end{tabular}
}
\end{table}

\vspace{-1em}
\subsection{Ablation Study}
To explore the effects of different parts of our method, we conduct several ablation experiments on the garment-adaptive pose predictor, the multi-scale appearance customization module, and the lightweight feature fusion.
\tabref{tab:ablation} shows the evaluation metrics of the ablation study. Removing the garment-adaptive pose predictor will greatly reduce the metrics as the generated pose maps encoded vital human structure information. For the appearance customization, the Showcase BLIP provides the overall information of the whole image. In the meantime, the Face BLIP introduces details to the results and equips GCO with the ability of fine-grained control. Without these modules, the metrics are decreased. Besides, to evaluate the effectiveness of the lightweight feature fusion, we change it with a commonly used IP-Adapter module to integrate the facial features. The resulting metrics verify that our carefully designed framework can achieve the best performance.  

\figref{fig:ablation}~(a) presents the results of the ablation study on our garment-adaptive pose predictor and the face fusion operation. The third column shows that the pose predictor is essential for maintaining the structure and proportions of the human body. In addition, the model with IP-Adapter may introduce unintended facial feature variations, \eg changing the natural lip color to red, whereas our methods can achieve better results.



\figref{fig:ablation}~(b) presents the results of the ablation study on the multi-scale appearance customization module. The second and third columns show results generated by the model without Showcase BLIP and the model without Face BLIP, respectively. The figure illustrates that the Showcase BLIP improves the image harmony and overall image characteristics, while the Face BLIP enables fine-grained customization of facial details.

\section{Conclusion and Limitation}

We present a novel garment-centric outpainting framework GCO to enable fine-grained controllable apparel showcase image generation. The framework takes a garment image segmented from a dressed mannequin or a person as input and generates apparel showcase images through a two-stage network. The first stage employs a garment-adaptive pose predictor to produce diverse pose maps conditioned on the garment image. The second stage integrates a lightweight feature fusion operation and a multi-scale appearance customization module for efficient image generation and fine-grained text-based control. Experiments demonstrate that GCO outperforms the state-of-the-art methods in realism and customizability. 
\vspace{-1em}
\paragraph{Limitation and Future Work.}
Since current fashion datasets contain data biases, such as the predominance of young female fashion models, the range of controllable attributes achievable by our method trained on these datasets is limited. Expanding these datasets to include a broader variety of ages, body types, and genders would be a promising direction for future work, as it would enhance the versatility and controllability of our method in generating more diverse and representative outputs.

{
    \small
    \bibliographystyle{ieeenat_fullname}
    \bibliography{main}
}


\end{document}